%% file: paper-template-latex/main.tex
\documentclass[letterpaper,10pt,journal,twoside]{IEEEtran}
\AtBeginDocument{\colorlet{yellow}{white}}
\usepackage{times}
\usepackage{amsmath}
\usepackage{amssymb}
\DeclareUnicodeCharacter{2191}{\ensuremath{\uparrow}}
\DeclareUnicodeCharacter{2193}{\ensuremath{\downarrow}}
\usepackage[table]{xcolor}
\usepackage{colortbl}
\usepackage{graphicx}
\usepackage{graphics}
\usepackage{epsfig}
\usepackage{booktabs}
\usepackage{multirow}
\usepackage{adjustbox}
\usepackage{siunitx}
\usepackage{mathtools}
\usepackage{wrapfig}
\usepackage{soul}
\DeclareRobustCommand{\added}[1]{\begingroup\sethlcolor{yellow}\hl{#1}\endgroup}
\usepackage{comment}
\usepackage{xspace}
\usepackage{etoolbox}
\usepackage[ruled,linesnumbered,noend]{algorithm2e}
\usepackage{cite}
\usepackage{hyperref}
\usepackage{stfloats}
\usepackage[percent]{overpic}
\usepackage[outline]{contour}
\usepackage[most]{tcolorbox}
\usepackage{array}
\usepackage{float}
\usepackage{placeins} %

\usepackage{xcolor}
\usepackage{pict2e}

\DeclareRobustCommand{\hatchsym}{%
  \begingroup
  \setlength{\unitlength}{1pt}%
  \raisebox{-0.20ex}{%
    \begin{picture}(14,7)
      \put(0,0){\color{black!05}\rule{14pt}{7pt}}%
      \linethickness{0.35pt}%
      \color{black!70}%
      \put(0,0){\framebox(14,7){}}%
      \put(0,0){\line(1,1){7}}%
      \put(3,0){\line(1,1){7}}%
      \put(7,0){\line(1,1){7}}%
      \put(11,0){\line(1,1){3}}%
      \put(0,4){\line(1,1){3}}%
    \end{picture}%
  }%
  \endgroup
}

\hypersetup{
    colorlinks=true,
    citecolor=black,
    linkcolor=black,
    filecolor=black,
    urlcolor=blue,
    pdftitle={Unified Perceptive Fall Safety for Humanoids},
    pdfauthor={Osher Azulay, Zhengjie Xu, Andrew Scheffer, Stella X. Yu},
    pdfsubject={},
    pdfkeywords={Humanoid Robot Systems, Whole-Body Motion Planning and Control, Visual Learning},
    pdfcreator={},
    pdfproducer={},
    pdfpagemode=FullScreen,
}

\newcommand{\best}[1]{\cellcolor{gray!10}\textbf{#1}}

\newcommand{\pmval}[1]{%
  \,\textcolor{gray!110}{\scalebox{0.65}{(\ensuremath{\pm}\,#1)}}%
}

\definecolor{filmcharcoal}{RGB}{100,100,90}
\definecolor{filmhole}{RGB}{238,235,228}
\newcommand{\filmrail}{%
  \begingroup\setlength{\fboxsep}{0pt}%
  \colorbox{filmcharcoal}{\hbox to\linewidth{%
    \rule{0pt}{0.85mm}%
    \leaders\hbox{\hspace{0.8mm}\textcolor{filmhole}{\rule[0.17mm]{0.5mm}{0.5mm}}\hspace{0.8mm}}\hfill}}%
  \endgroup
}
\newcommand{\filmframe}[3][0pt 0pt 0pt 0pt]{%
  \begin{minipage}[b]{#2}%
    \setlength{\parskip}{0pt}\offinterlineskip
    \filmrail\par\nointerlineskip
    \includegraphics[width=\linewidth,trim=#1,clip]{#3}\par\nointerlineskip
    \filmrail
  \end{minipage}%
}

\definecolor{imagelabelbackground}{HTML}{292B2C}
\newcommand{\imagelabel}[1]{%
  \tikz[baseline=(label.base)]{%
    \node[fill=imagelabelbackground,fill opacity=.73,text=white,text opacity=1,draw=none,
          rounded corners=.5pt,inner xsep=1.4pt,inner ysep=1pt]
      (label) {#1};%
  }%
}

\newcommand{\algo}{\textsc{VIGOR}\xspace}
\def\sname{VIGOR}

\title{Unified Perceptive Fall Safety for Humanoids}

\author{
Osher Azulay, Zhengjie Xu, Andrew Scheffer, and Stella X. Yu%
\thanks{Manuscript received May 5, 2026; revised October 2, 2026; accepted October 13, 2026.}
\thanks{This paper was recommended for publication by Editor Olivier Stasse upon evaluation of the Associate Editor and Reviewers' comments.}
\thanks{All authors are with the University of Michigan, Ann Arbor, MI, USA. 
Corresponding Author: {\tt\footnotesize osher@umich.edu}.}
\thanks{Digital Object Identifier (DOI): see top of this page.}
}
\begin{document}

\setlength{\belowdisplayskip}{2pt}
\setlength{\belowdisplayshortskip}{3pt}
\setlength{\abovedisplayskip}{2pt}
\setlength{\abovedisplayshortskip}{3pt}
\setlength{\parskip}{0pt}
\raggedbottom

\markboth{IEEE Robotics and Automation Letters. Preprint Version. Accepted October, 2026}
{Azulay \MakeLowercase{\textit{et al.}}: Unified Perceptive Fall Safety for Humanoids}

\maketitle

\input{paper-template-latex/0_abstract}

\section{Introduction}

\input{paper-template-latex/1_intro_new}

\section{Related Work}
\label{sec:related}
\input{paper-template-latex/2_related}

\section{Method}
\input{paper-template-latex/3_method}

\section{Experiments}
\input{paper-template-latex/4_experiments}


\FloatBarrier
\section{Conclusion}
\label{sec:conclusion}
\input{paper-template-latex/5_conclusions}

\bibliographystyle{IEEEtran}
\bibliography{paper-template-latex/references}


\end{document}

%% file: paper-template-latex/0_abstract.tex
\begin{abstract}

Reliable fall recovery is critical for humanoids operating in cluttered environments. Unlike quadrupeds or wheeled robots, humanoids undergo high-energy impacts, complex whole-body contact, and large viewpoint changes during falls, making recovery essential.
Prior work treats fall safety as separate problems (avoidance, impact mitigation, stand-up) or learns end-to-end policies without vision, typically on flat terrain. More fundamentally, it couples pose, dynamics, and terrain into a single data space, requiring exhaustive coverage and limiting generalization.
We present a unified visually grounded fall-safety approach spanning all phases of recovery. It builds on two insights: (1) human fall-recovery poses are structured and transferable across terrains via alignment, and (2) \added{effective recovery depends on interpreting the recovery goal in relation to the surrounding terrain}. We train a privileged teacher using sparse human demonstrations and simulated terrains, and distill it into a deployable student that relies on egocentric depth and proprioception. \added{The student predicts a goal-in-context latent representing the recovery target in relation to its surroundings, guiding coordinated whole-body reactions.} Experiments in simulation and on a Unitree G1 demonstrate robust zero-shot fall safety across diverse terrains.
Project page: \href{https://vigor2026.github.io/}{vigor2026.github.io}.

\end{abstract}

%% file: paper-template-latex/1_intro_new.tex
\IEEEPARstart{H}{umanoids} are intended to operate in the same cluttered and uneven environments as humans, yet they remain vulnerable when failures occur. Minor disturbances can quickly cascade into coupled rotations, impacts, and contact transitions, unfolding faster than conventional stabilization can correct. \added{Humanoid fall safety demands whole-body coordination under severe time pressure, as early actions constrain later contacts and recovery.}

Fall safety cannot be reduced to a post-impact reflex: How a robot falls determines which supports and contacts remain feasible, and whether standing up is possible at all. These choices depend on situational and terrain awareness, since contact feasibility and momentum redirection are shaped by local geometry invisible to proprioception alone. Fall safety thus requires the integration of terrain perception and coordinated whole-body control across the fall-recovery process. Despite the intrinsic coupling between fall and recovery, existing approaches fragment fall safety along two axes and typically rely on blind, proprioceptive-only sensing.

\begin{figure}[t]
\centering
\setlength{\fboxsep}{0pt}
\setlength{\fboxrule}{0.3pt}

\newlength{\gap}
\setlength{\gap}{1.5pt}

\newlength{\leftW}
\setlength{\leftW}{0.60\columnwidth}
\newlength{\rightW}
\setlength{\rightW}{\dimexpr\columnwidth-\leftW-\gap\relax}

\newlength{\teaserH}
\settoheight{\teaserH}{\includegraphics[width=\leftW]{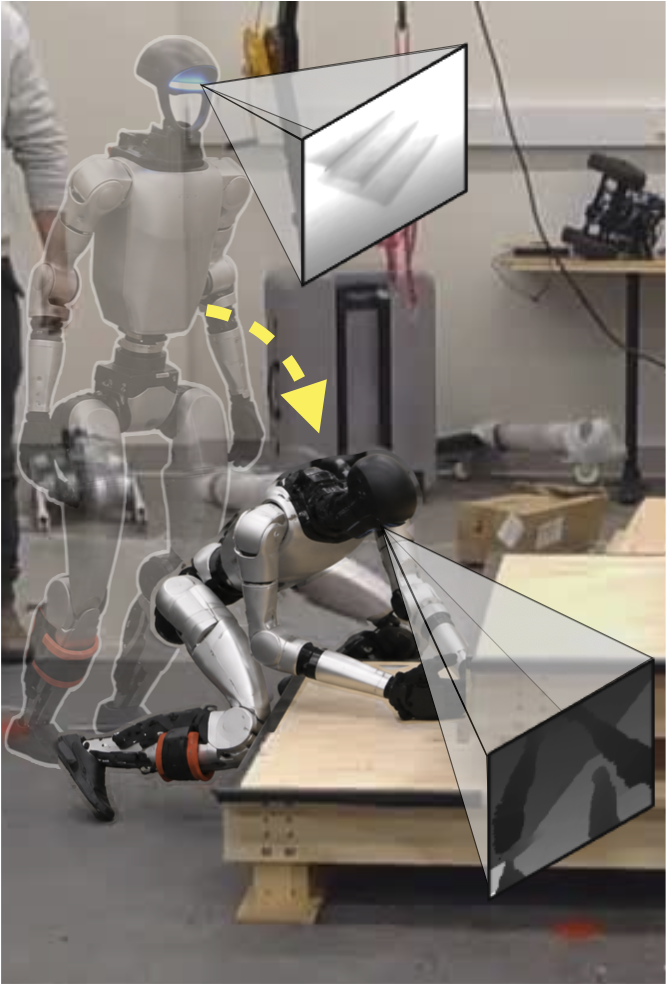}}

\newlength{\smallH}
\setlength{\smallH}{\dimexpr(\teaserH - 2\gap - 4\fboxrule)/3\relax}

\begin{minipage}[t]{\leftW}
\vspace{0pt}
\fcolorbox{black}{white}{%
\includegraphics[width=\leftW]{figures/teaser/diversity/big_first_page.png}}
\end{minipage}%
\hspace{\gap}%
\begin{minipage}[t]{\rightW}
\vspace{0pt}
\fcolorbox{black}{white}{\includegraphics[width=\rightW,height=\smallH,keepaspectratio]{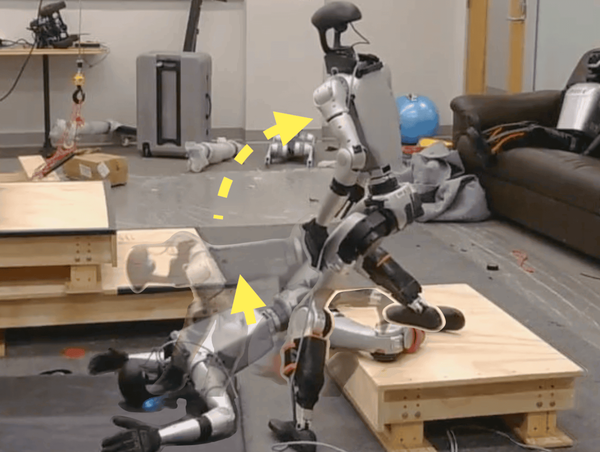}}

\vspace{\gap}

\fcolorbox{black}{white}{\includegraphics[width=\rightW,height=\smallH,keepaspectratio]{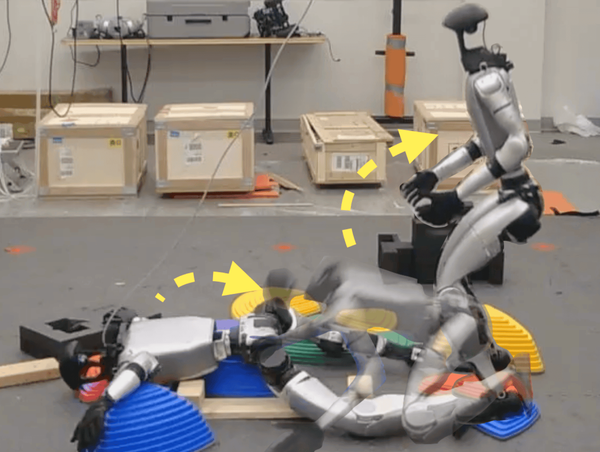}}

\vspace{\gap}

\fcolorbox{black}{white}{\includegraphics[width=\rightW,height=\smallH,keepaspectratio]{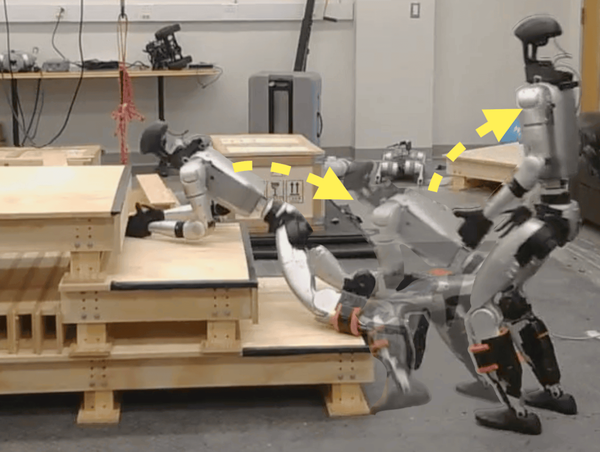}}
\end{minipage}

\caption{\textbf{Vision-enabled unified fall safety for humanoids}.
Left: fall-mitigation recovery after a push towards stairs demands vision-guided anticipatory arm contact to arrest the fall.
Right: recovery behaviors across diverse terrain and initial conditions highlight the generality of vision-guided contact selection and whole-body rebalancing.}
\label{fig:teaser}
\vspace{-5mm}
\end{figure}

First, many methods decompose fall-related behaviour into isolated subproblems: \added{fall avoidance (preventing loss of balance), impact mitigation (controlling descent and contact to reduce impact), or stand-up recovery (restoring stable standing).} Balance controllers focus on preventing loss of stability~\cite{englsberger2014trajectory, ferigo2021emergence}, while stand-up controllers are typically invoked only after the robot reaches rest in a small set of predefined poses~\cite{luo2014multi}. These components are designed largely in isolation and rely primarily on proprioceptive sensing, implicitly assuming flat ground and benign contact geometry. Recovery therefore cannot be separated from \textit{how the robot falls}, nor from \textit{where and how contact occurs}.

Second, learning-based approaches that aim to address fall-related behaviors end-to-end often treat the problem as one of monolithic data complexity, and are typically trained without access to visual terrain perception.  Reinforcement learning (RL) methods usually require extensive reward engineering, frequently leading to unnatural motions~\cite{HuangT-RSS-25}. Imitation learning methods rely on dense trajectory demonstrations, often sourced from internet-scale human motion data, which transfer poorly across terrain geometry and contact conditions~\cite{zargarbashi2024robotkeyframing, wen2025constrained, zhang2025motion}. Both families of methods often treat environmental complexity as a monolithic data problem, entangling kinematics, dynamics, and terrain and requiring exhaustive coverage of their combinations.

\begin{figure*}[!t]
\centering
\includegraphics[width=0.99\textwidth]{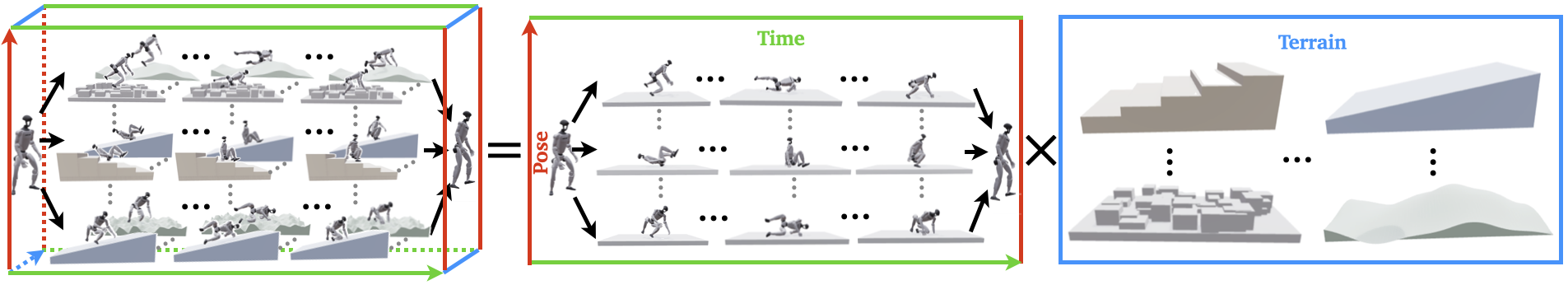}
\caption{\textbf{Factorized data generation yields sample-efficient imitation and scalable adaptation for humanoid fall safety learning.}
Rather than treating \textcolor{red}{pose}, \textcolor{green}{time}, and \textcolor{blue}{terrain} as a single monolithic data space requiring exhaustive coverage, we factorize it into a small set of human pose trajectories from real-world demonstrations on flat terrain and independently varying terrain geometry in simulation.}
\label{fig:FacterizeData}
\vspace{-3mm}
\end{figure*}

In contrast, we adopt a factorized view of data complexity (Fig.~\ref{fig:FacterizeData}): Our first insight is that natural human fall and recovery poses are far more constrained than they appear: Poses required on complex terrain often admit spatially aligned and physically compatible counterparts on flat terrain.  This observation allows pose and terrain variation to be treated as largely independent factors. As a result, a few human demonstrations collected on flat ground can serve as priors on kinematics and dynamics, while variation in terrain geometry, contact timing, and momentum redirection is handled through RL via interaction with the environment.

\added{Our second insight is that recovery goals must be understood in relation to the terrain on which they are realized. A motion prior provides whole-body coordination, while the current body state and local geometry determine which contacts can support that recovery.}

We therefore build the policy directly on a compact visual goal-in-context latent representation, which integrates the next target pose with the local terrain and body context in a single perceptual-motor space.

Based on these insights, we introduce \sname\ ({\it \underline{Vi}sual \underline{Go}al-In-Context Infe\underline{r}ence}), \added{a perceptive control framework that unifies fall avoidance, impact mitigation, and stand-up recovery.}  A privileged teacher policy is trained via RL using sparse human demonstrations on flat terrain together with access to local terrain geometry, yielding terrain-aware reactive behaviours. This knowledge is then distilled into a deployable student policy that operates using only egocentric depth observations and short-term proprioceptive history. \added{Latent supervision guides the student's interpretation of its recovery target, while action imitation guides the resulting control.}

Our contributions are threefold:
\begin{itemize}
    \item We formulate humanoid fall safety as a unified, \textit{visually grounded} control problem spanning fall mitigation, impact handling, and stand-up recovery within a single policy. \added{To our knowledge, this is the first learning-based framework for unified perceptive humanoid recovery.}

    \item We introduce a factorized data formulation that decouples human fall-recovery motion from terrain variation, enabling sample-efficient learning from a small set of flat-ground human demonstrations combined with large-scale terrain randomization.

    \item \added{We introduce a visual recovery policy trained through intermediate latent supervision alongside action imitation. A privileged teacher guides the student to infer recovery targets suited to the surrounding terrain, using only egocentric depth and proprioceptive history.}
\end{itemize}

We evaluate \sname\ extensively in simulation and demonstrate zero-shot transfer to a real humanoid across diverse fall and recovery conditions, validating a unified vision-enabled approach to humanoid fall safety.

%% file: paper-template-latex/2_related.tex
\begin{figure*}[!t]
\centering
\includegraphics[width=0.99\textwidth]{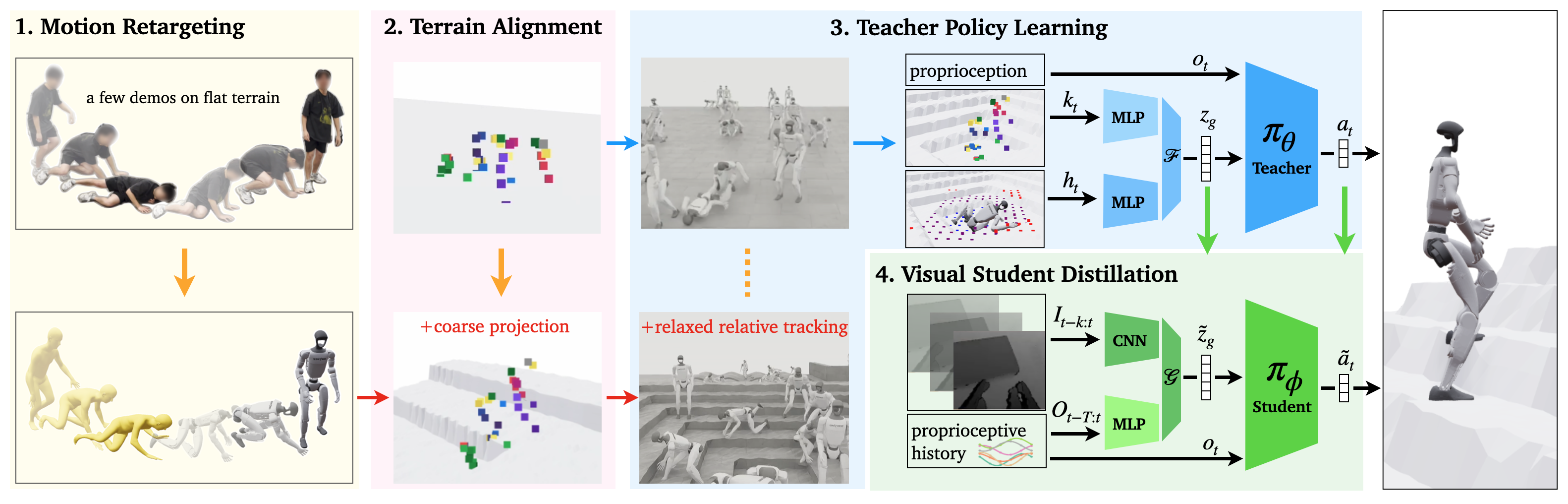}
\vspace{-2mm}
\caption{
\textbf{Overview of \algo.}
{\bf 1)} Human fall--recovery demonstrations \added{collected on flat ground} are kinematically retargeted to the robot.
{\bf 2)} Reference poses are \added{used directly} on flat terrain and coarsely projected onto uneven terrain to provide sparse tracking targets.
{\bf 3)} A privileged RL teacher learns a goal-in-context latent \added{representing the immediate recovery target in relation to local terrain}.
{\bf 4)} \added{A student learns through latent supervision and action imitation, using only egocentric depth and proprioceptive history at deployment.}
}
\label{fig:Workflow}
\vspace{-5mm}
\end{figure*}

\subsection{Humanoid Fall Mitigation and Standing-Up Control}

\added{Humanoid fall safety requires managing impacts and restoring stable standing.} Prior work typically decomposes recovery into two disjoint stages: fall mitigation during descent and standing up after impact. Classical fall mitigation methods focus on maintaining balance or redirecting momentum to reduce impact forces~\cite{ferigo2021emergence,englsberger2014trajectory}, while learning-based approaches similarly emphasize safe landing or damage reduction without explicitly addressing subsequent recovery~\cite{kumar2017learning}. \added{Stand-up controllers typically act after the robot has settled into predefined configurations}~\cite{stuckler2006getting}, \added{although recent learning-based methods extend recovery to more diverse initial poses}~\cite{HuangT-RSS-25,HeX-RSS-25}. Only a few works attempt to unify fall mitigation and standing-up within a single policy~\cite{gaspard2025frasa,xu2025unified}, but these approaches typically rely on limited sensing or blind operation. In contrast, we learn a \textit{unified, visually grounded} policy that explicitly couples fall mitigation and recovery under diverse terrain conditions.

\subsection{Visual Whole-Body Control}

Methods in vision-based control have recently enabled humanoids to perform complex whole-body behaviors under partial observability \cite{zhuang2024humanoid}. Visual input has been used to support loco-manipulation and contact-rich interaction~\cite{fu2022deep,liu2024visual,yin2025visualmimic}, as well as to guide navigation and locomotion across challenging environments~\cite{loquercio2023learning,lin2025let,duan2023learning}. Visual reinforcement learning has also produced active perception during whole-body manipulation in simulated humanoids~\cite{luo2025emergent}. Despite this progress, fall recovery remains perceptually challenging: rapid body rotations, self-occlusion, and intermittent ground contact produce unstable visual observations. Visually grounded fall recovery is still largely unexplored, especially for policies spanning fall mitigation, impact, and stand-up recovery.

\subsection{Motion Priors and Style-Constrained RL}
\added{Learning whole-body control requires coordinating many joints through changing contacts.} Prior work therefore relies on carefully shaped multi-term reward functions~\cite{wang2025beamdojo, zhuang2024humanoid} or constrains the control problem through reduced body representations or limited link actuation~\cite{lin2025let}. Human motion priors provide an alternative form of regularization, inducing behavior with minimal reward engineering.

Reference-guided RL methods such as DeepMimic~\cite{peng2018deepmimic} and AMP~\cite{peng2021amp} demonstrate strong whole-body control on flat terrain, but often rely on dense trajectory tracking or periodic motion assumptions, which become limiting in highly non-periodic, contact-rich behaviors such as fall recovery. Recent work relaxes dense tracking through higher-level structural guidance, including keyframe-based objectives~\cite{zargarbashi2024robotkeyframing}, style-constrained learning~\cite{wen2025constrained}, or hybrid formulations that combine flat-terrain imitation with residual adaptation on uneven terrain~\cite{zhang2025motion}. These methods allow deviation from demonstrations when needed, but still depend on sufficient motion--terrain coverage. We therefore treat human fall-recovery demonstrations as \emph{sparse structural priors} rather than dense trajectory targets.

%% file: paper-template-latex/3_method.tex

We propose \algo, a unified framework for humanoid stand-up, fall mitigation, and recovery in unstructured environments, as illustrated in Fig.~\ref{fig:Workflow}. Unlike pipelines that treat falling, impact mitigation, and standing as separate modules, \algo learns the full process with a single control policy conditioned on onboard egocentric sensing.

Following a destabilizing disturbance, the robot may enter arbitrary configurations involving high-energy impacts, whole-body contact, and large changes in body orientation. At each timestep $t$, the robot receives proprioception
$
\mathbf{o}_t =
[\boldsymbol{\omega}_t,\mathbf{g}_t,\mathbf{q}_t,\dot{\mathbf{q}}_t,\mathbf{a}_{t-1}],
$
where $\boldsymbol{\omega}_t$ is the base angular velocity, $\mathbf{g}_t$ is projected gravity, $\mathbf{q}_t$ and $\dot{\mathbf{q}}_t$ are joint positions and velocities, and $\mathbf{a}_{t-1}$ is the previous action. The policy outputs joint-space PD targets $\mathbf{a}_t$ for the 23-DoF humanoid.

To learn recovery under these conditions, we adopt a privileged teacher--student design. During training, the teacher $\pi_{\theta}$ observes the current proprioception $\mathbf{o}_t$, sparse keyframe targets $\mathbf{k}_t$, and local terrain samples $\mathbf{h}_t$, which provide structured recovery intent and terrain context. The deployable student $\pi_{\phi}$ does not access these privileged inputs. Instead, it receives proprioceptive history $\mathbf{o}_{t-T:t}$ and stacked egocentric depth $\mathbf{I}_{t-k:t}$ from a head-mounted camera, and learns to reconstruct the teacher's \emph{visual goal-in-context} latent. This enables unified vision-based fall mitigation and recovery from onboard sensing.

\subsection{\added{Motion--Terrain Factorization}}
\label{sec:motion_collection}

\added{Motion--terrain factorization removes the need for terrain-matched demonstrations: flat-ground motions supply recovery structure, while RL learns terrain-specific contacts and dynamics.} We obtain fall--recovery demonstrations from monocular human videos recorded on flat ground, covering prone, forward, sideways, and backward recoveries. We process the videos with VideoMimic~\cite{allshire2025visual} and retarget the reconstructed motions to the Unitree~G1 using kinematics-aware mapping with conservative pelvis and hip constraints to reduce artifacts. We retain nine high-quality recovery sequences from three body sizes.
Rather than tracking full trajectories, we uniformly subsample sparse keyframes at $\sim$5\,Hz. These keyframes provide coarse structural priors for the privileged teacher while allowing RL to adapt contact timing, momentum redirection, and terrain-dependent recovery.
During training, keyframes are coarsely aligned to terrain by a vertical shift:
\[
\Delta z =
\max_{i=1,\dots,N_{\text{links}}}
\bigl(h(\mathbf{p}^{\text{ref}}_{i}) - z^{\text{ref}}_{i}\bigr),
\]
where $h(\cdot)$ is the terrain height at link position $\mathbf{p}^{\text{ref}}_{i}$. All reference poses are shifted by $\Delta z$ to maintain terrain clearance and are initialized near the terrain center with small $x$--$y$ perturbations. To increase coverage, we randomize the reference trajectory, temporal phase, global transformations, and occasionally add a short free-fall interval by suppressing a random subset of joint torques before control begins.

\subsection{Privileged Goal-in-Context Teacher}
\label{sec:teacher}
\vspace{-1mm}

We train a privileged fall-mitigation and recovery teacher policy $\pi_{\theta}$ using PPO~\cite{ppo}. Training proceeds in two stages: first on flat terrain to acquire basic fall mitigation and stand-up behaviors, and then on randomized non-flat terrains to handle diverse contact geometry and perturbations. The teacher receives proprioception $\mathbf{o}_t$, sparse keyframe targets $\mathbf{k}_t$, and privileged terrain samples $\mathbf{h}_t$. The reference and terrain signals are fused into a single \emph{goal-in-context} latent
$
\mathbf{z}_g
=
\mathcal{F}\bigl(\mathbf{k}_t,\,\mathbf{h}_t\bigr),
$
which encodes the immediate recovery target conditioned on local terrain context. \added{We normalize the projected latent as} \colorbox{yellow}{$\mathbf{z}_g=\sqrt{d}\,\bar{\mathbf{z}}_g/\|\bar{\mathbf{z}}_g\|_2$} \added{to maintain a consistent scale for policy conditioning and distillation}. The teacher actor conditions on this latent and proprioception to produce joint targets:
$
\mathbf{a}_t
=
\pi_{\theta}
\bigl(
\mathbf{z}_g,\,
\mathbf{o}_t
\bigr).
$
While the teacher is trained in a tracking-based manner similar to prior motion imitation policies~\cite{peng2018deepmimic}, the reference signals are intentionally sparse and enforced only at a coarse, relative level. Together with privileged terrain observations, this under-specification allows RL to resolve contact timing, body placement, and terrain-dependent execution details beyond what is prescribed by the demonstrations.


\subsection{Reward Design}
We design a composite reward function to support fall mitigation, stabilization, and recovery within a single policy:
\[
r_t = r_t^{\text{imit}} + r_t^{\text{reg}} + r_t^{\text{post}}
\]
where $r_t^{\text{imit}}$ is a DeepMimic-style tracking term~\cite{peng2018deepmimic}$,$
$r_t^{\text{reg}}$ aggregates standard motion-regularization penalties,
and $r_t^{\text{post}}$ provides post-recovery stabilization by rewarding upright hold and suppressing residual motion once standing. All reward terms are summarized in Table~\ref{tab:reward_summary}.

\textbf{Motion imitation.}
We project reference poses, recorded on flat ground, onto the simulated terrain before computing tracking errors, as described in Sec.~\ref{sec:motion_collection}. This preserves the demonstrated motion structure but is not physically exact on uneven terrain, leaving small residual offsets. Tracking full world-frame positions $\|\mathbf{p}^{\text{ref}}_{i,t} - \mathbf{p}_{i,t}\|$ would therefore penalize projection residuals and terrain-dependent deviations needed for recovery. Instead, we define imitation targets in a root-relative canonical frame by comparing $(\mathbf{p}^{\text{ref}}_{i,t} - \mathbf{p}^{\text{ref}}_{0,t}) - (\mathbf{p}_{i,t} - \mathbf{p}_{0,t})$, preserving relative body configuration while remaining insensitive to projection artifacts, allowing demonstrations to act as structural priors rather than exact pose targets.

\textbf{Regularization and safety.}
The term $r_t^{\text{reg}}$ aggregates standard regularization penalties, including joint limit violations, joint velocities, accelerations, momentum change, action smoothness, undesired contacts, and contacts near terrain edges. These terms stabilize learning and reduce unsafe behaviors without prescribing a specific recovery strategy.

\textbf{Post-recovery stabilization.}
Once the robot reaches an upright configuration, $r_t^{\text{post}}$ encourages stable upright posture by suppressing residual base motion and joint velocities.

\begin{table}[tbp]
\centering
\begingroup
\scriptsize
\setlength{\tabcolsep}{3pt}
\renewcommand{\arraystretch}{0.90}
\setlength{\aboverulesep}{0.3ex}
\setlength{\belowrulesep}{0.3ex}
\caption{\textbf{Reward terms used for fall-recovery learning}.}
\vspace{-1mm}
\label{tab:reward_summary}
\setlength{\fboxsep}{1pt}
\setlength{\fboxrule}{0.8pt}
\begin{tabular}{l|c|c}
\toprule
\textbf{Reward} & \textbf{Definition} & \textbf{Weight} \\
\midrule
\multicolumn{3}{c}{\textit{Motion imitation}} \\
\midrule
Pos. track.        & $\exp(-\|\Delta x\|^2 / \sigma)$            & \scalebox{0.70}{$1.25$} \\
Rot. track.        & $\exp(-\|\Delta \theta\|^2 / \sigma)$       & \scalebox{0.70}{$0.5$} \\
Lin. vel. track.   & $\exp(-\|\Delta v\|^2 / \sigma)$            & \scalebox{0.70}{$0.125$} \\
Ang. vel. track.   & $\exp(-\|\Delta \omega\|^2 / \sigma)$       & \scalebox{0.70}{$0.125$} \\
Joint pos. track.  & $\exp(-\|\Delta q\|^2 / \sigma)$            & \scalebox{0.70}{$0.5$} \\
Joint vel. track.  & $\exp(-\|\Delta \dot q\|^2 / \sigma)$       & \scalebox{0.70}{$0.125$} \\
\midrule
\multicolumn{3}{c}{\textit{Regularization \& safety}} \\
\midrule
Torque penalty     & $\|\tau\|^2$                                & \scalebox{0.70}{$-10^{-6}$} \\
Torque limit       & $\max(0, |\tau|-\tau_{\max})$               & \scalebox{0.70}{$-0.1$} \\
Joint pos. limit   & $\max(0, q-q_{\max})+\max(0, q_{\min}-q)$   & \scalebox{0.70}{$-10$} \\
Joint vel. limit   & $\max(0, |\dot q|-\dot q_{\max})$           & \scalebox{0.70}{$-5$} \\
Joint acc.         & $\|\ddot q\|^2$                             & \scalebox{0.70}{$-2.5{\cdot}10^{-7}$} \\
Momentum change    & $\|\Delta p\|$                              & \scalebox{0.70}{$-5{\cdot}10^{-3}$} \\
Body yank          & $\|\Delta F\|^2$                            & \scalebox{0.70}{$-2{\cdot}10^{-6}$} \\
Joint vel. penalty & $\|\dot q\|^2$                              & \scalebox{0.70}{$-10^{-4}$} \\
Action rate        & $\|a_t-a_{t-1}\|^2$                         & \scalebox{0.70}{$-0.1$} \\
Undesired contacts & $r_{\text{contacts}}$ \cite{liao2025beyondmimic} & \scalebox{0.70}{$-0.1$} \\
Foothold           & $r_{\text{foothold}}$ \cite{wang2025beamdojo} & \scalebox{0.70}{$-1$} \\
\midrule
\multicolumn{3}{c}{\textit{Post-recovery stabilization}} \\
\midrule
Head height        & $\exp(-\|\Delta h\|^2 / \sigma)$            & \scalebox{0.70}{$0.25$} \\
Base lin. vel.     & $\|\mathbf{v}_{\text{base}}\|^2$            & \scalebox{0.70}{$-0.01$} \\
Base ang. vel.     & $\|\boldsymbol{\omega}_{\text{base}}\|^2$   & \scalebox{0.70}{$-0.05$} \\
\bottomrule
\end{tabular}%
\endgroup
\vspace{-4mm}
\end{table}

\subsection{Egocentric Student Policy}

At deployment time, the robot relies only on onboard sensing. The student therefore learns to infer the teacher's goal-in-context latent from egocentric depth and short-term proprioceptive history. The same sparse demonstrations used to train the teacher define the space of recovery behaviors available to the student, which infers an appropriate recovery target without explicit reference signals.

The student receives egocentric depth $\mathbf{I}_{t-k:t}$ and proprioceptive history $\mathbf{o}_{t-T:t}$. These inputs are fused into a predicted \textit{visual} goal-in-context latent
$
\tilde{\mathbf{z}}_g
=
\mathcal{G}\bigl(\mathbf{I}_{t-k:t},\,\mathbf{o}_{t-T:t}\bigr),
$
which is designed to capture the terrain-conditioned recovery objective. \added{To infer this latent, the student encodes depth with a CNN and proprioceptive history with a temporal encoder. The resulting features are concatenated and projected through an MLP, then normalized as in the teacher.} The student actor conditions on this latent and current proprioception to produce joint targets:
$
\tilde{\mathbf{a}}_t
=
\pi_{\phi}
\bigl(
\tilde{\mathbf{z}}_g,\,
\mathbf{o}_t
\bigr).
$

We train the student by distilling from the privileged teacher using a hybrid of behavioral cloning and DAgger-style~\cite{dagger, azulay2025visuotactile} rollouts. Supervision is applied over a mixture of teacher- and student-induced state distributions,
\[
\rho
=
\alpha \rho_{\pi_{\theta}}
+
(1-\alpha)\rho_{\pi_{\phi}},
\]
where $\rho_{\pi_{\theta}}$ and $\rho_{\pi_{\phi}}$ denote the distributions induced by teacher and student rollouts, respectively. The mixing coefficient $\alpha$ is annealed from $1$ toward $0$ during training, gradually shifting rollouts from teacher-driven supervision to student-induced states. The student is trained using weighted behavior-cloning and goal-in-context reconstruction losses:
\[
\mathcal{L}_{\text{distill}}
=
\mathbb{E}_{t\sim \rho}
\left[
\lambda_{z}\,\bigl\|\tilde{\mathbf{z}}_g - \mathbf{z}_g\bigr\|^2
+
\lambda_{a}\,\bigl\|\tilde{\mathbf{a}}_t - \mathbf{a}_t\bigr\|^2
\right].
\]

Teacher rollouts provide stable demonstrations that rapidly imprint recovery behavior, while student rollouts expose the policy to off-distribution states and improve adaptability.  

Although the teacher is trained with motion-tracking signals that include intentional falls, the student is distilled from unstable initial states, which biases it toward stabilization rather than deliberate falling. Therefore, without disturbance the policy maintains a stable posture; under disturbance, it mitigates fall severity and recovers.


\subsection{Domain Randomization}

To improve robustness and sim-to-real transfer, following prior works~\cite{azulay2025visuotactile, yin2025visualmimic}, we randomize both dynamics and perception at the start of each episode and throughout rollout. On the dynamics side, we randomize friction and restitution, and further vary the episode initialization described in Sec.~\ref{sec:motion_collection}, including the reference motion clip and phase, planar position, yaw, initial pose, and joint states. We additionally apply stochastic external pushes and occasional joint torque dropouts to mimic disturbances and partial actuator failures.
On the perception side, we perturb depth observations using depth clipping and non-linear remapping, multiplicative noise, spatial and temporal dropout, synthetic occlusions, and small camera pose jitter. These perturbations encourage the policy to rely on stable geometric structure.

%% file: paper-template-latex/4_experiments.tex
\begin{figure}[t]
\centering
\setlength{\abovecaptionskip}{1pt}
\begingroup
\setlength{\fboxsep}{1pt}
\fcolorbox{yellow}{white}{\includegraphics[width=\dimexpr\linewidth-2\fboxsep-2\fboxrule\relax]{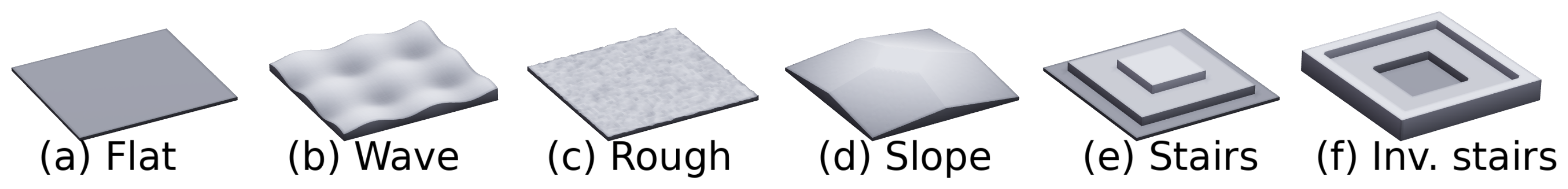}}
\endgroup
\vspace{-4mm}
\caption{
\added{Training terrains, from left to right: flat, wave, rough, slope, stairs, and inverted stairs.} One representative difficulty is shown.
}
\vspace{-4mm}
\label{fig:terrains}
\end{figure}

\subsection{Implementation Details}

We summarize the main components of our simulation, training, and deployment setup below. For more visual results, see the supplementary multimedia.

\textbf{Simulation Setup.}
We use a 23-DoF Unitree G1 with a head-mounted depth camera in IsaacGym and IsaacSim, controlled at 50 Hz with a 7.5 s recovery horizon.
Training proceeds in two phases. The policy is first trained on flat terrain to acquire core stand-up and impact-mitigation behaviors, and is then continued on randomized non-flat terrains, as shown in Fig.~\ref{fig:terrains}.
We train the teacher with PPO using 4,096 parallel environments, then distill it into the visual student using 512 environments.

\textbf{Real-World Setup:}
Hardware experiments are performed on a Unitree~G1 equipped with an Intel RealSense head mounted depth camera. Proprioception streams at $500$\,Hz and depth at $30$\,Hz. The student policy runs at \texttt{50}\,Hz and outputs joint-space position targets to the low-level PD controller. Depth preprocessing mirrors simulation, and deployment is \textit{fully zero-shot} without any real-world fine-tuning.

\subsection{Simulated Experiments}

We evaluate policies under two initialization regimes on terrains sampled from the families shown in Fig.~\ref{fig:terrains}. In both cases, initial planar position and yaw are randomized.

In \textit{Stand-Up}, episodes start from low-body configurations sampled around the lowest keyframe of each demonstration, with pose noise. In \textit{Fall-Recovery}, episodes start earlier in the motion, near the onset of falling, with random perturbations and nonzero base velocity, requiring the policy to handle impact mitigation and recovery on varying terrain geometry. Unless otherwise stated, results are reported in IsaacGym.

\textbf{Metrics.} \added{Simulation results in this section use 500 trials per seed across three different evaluation seeds. Table}~\ref{tab:metrics} \added{defines the evaluation metrics.}

\begin{table}[t]
\caption{\added{Evaluation metrics.}}
\label{tab:metrics}
\vspace{-2mm}
\centering
\begingroup
\footnotesize
\setlength{\fboxsep}{1pt}
\setlength{\fboxrule}{0.8pt}
\renewcommand{\arraystretch}{1.08}
\fcolorbox{yellow}{white}{%
\begin{tabular}{@{}>{\raggedright\arraybackslash}p{0.21\linewidth}@{\hspace{5pt}}>{\raggedright\arraybackslash}p{0.75\linewidth}@{}}
\toprule
\textbf{Metric} & \textbf{Definition} \\
\midrule
SR $\uparrow$ (\%) & Success with upright stabilization within $7.5$ s. \\
SSR $\uparrow$ (\%) & Stable Success without re-fall. \\
Time $\downarrow$ (s) & Time to recovery. \\
PIF $\downarrow$ & Peak internal-force magnitude. \\
$F_{\mathrm{c}}$ $\downarrow$ (kN) & Peak contact force across all links. \\
FR $\downarrow$ (\%) & Subsequent re-fall rate. \\
Drift $\downarrow$ (m) & Horizontal pelvis displacement. \\
MPJPE $\downarrow$ (cm) & Pelvis-relative joint tracking error. \\
\bottomrule
\end{tabular}%
}
\endgroup
\end{table}

\subsection{Simulated Results}

\textbf{Teacher Ablations.}
We evaluate teacher-side ablations to isolate the contribution of privileged structure and supervision. NoKeypoints removes keypoint-based observations. DofKeypoints replaces spatial keypoint positions with joint angle targets. AbsTrack trains the teacher using absolute pose tracking objectives instead of relative tracking. NoScandots removes access to privileged terrain information, while Teacher denotes the full privileged teacher.

\begin{table}[t]
\setlength{\tabcolsep}{1.8pt}
\caption{Teacher ablations under Stand-Up and Fall-Recovery.}%
\label{tab:TeacherAblate}
\vspace{-2mm}
\centering

\begingroup
\setlength{\fboxsep}{1pt}
\setlength{\fboxrule}{0.8pt}
\fcolorbox{yellow}{white}{%
\begin{tabular}{@{}l ccc@{\hspace{6pt}}ccc@{}}
\toprule
\multirow{2}{*}{\textbf{Method}}
& \multicolumn{3}{c}{\textbf{Stand-Up}}
& \multicolumn{3}{c}{\textbf{Fall-Recovery}} \\
\cmidrule(lr){2-4}\cmidrule(l){5-7}
& SR↑ & SSR↑ & Time↓ & SR↑ & SSR↑ & Time↓ \\
\midrule
NoKeypoints&
57.4\pmval{4.1} & 53.2\pmval{3.0} & 6.9\pmval{1.3} &
76.1\pmval{3.1} & 73.2\pmval{3.4} & 6.3\pmval{1.5} \\
DofKeypoints&
86.6\pmval{3.1} & 80.5\pmval{3.4} & 5.8\pmval{1.0} &
84.1\pmval{2.9} & 80.1\pmval{2.4} & 5.5\pmval{0.9} \\
AbsTrack&
73.1\pmval{3.0} & 70.0\pmval{3.5} & 5.7\pmval{2.0} &
70.4\pmval{3.5} & 67.3\pmval{3.8} & 5.6\pmval{2.1} \\
NoScandots&
84.4\pmval{3.8} & 70.0\pmval{3.4} & 5.6\pmval{1.7} &
88.3\pmval{3.7} & 77.6\pmval{4.6} & 6.4\pmval{1.6} \\
Teacher&
\best{98.0\pmval{0.3}} & \best{97.6\pmval{0.2}} & \best{4.2\pmval{1.0}} &
\best{96.3\pmval{0.8}} & \best{95.7\pmval{1.2}} & \best{5.3\pmval{0.9}} \\
\bottomrule
\end{tabular}%
}
\endgroup
\vspace{-3mm}
\end{table}

Table~\ref{tab:TeacherAblate} highlights the role of structural priors in the teacher. \added{Removing keypoint guidance causes the largest degradation in Stand-Up, while replacing relative tracking with absolute tracking causes the largest degradation in Fall-Recovery.}
Similarly, replacing spatial keypoints with joint-angle targets (DofKeypoints) partially recovers performance but remains below the full model, suggesting that geometric body configuration provides stronger guidance than joints alone. Removing privileged terrain samples (NoScandots) limits the teacher's scene awareness. The full privileged Teacher achieves the best overall performance.

\begin{figure*}[t]
\centering
\begingroup
\setlength{\fboxsep}{1pt}
\setlength{\fboxrule}{0.8pt}
\fcolorbox{yellow}{white}{\includegraphics[width=\dimexpr\linewidth-2\fboxsep-2\fboxrule\relax]{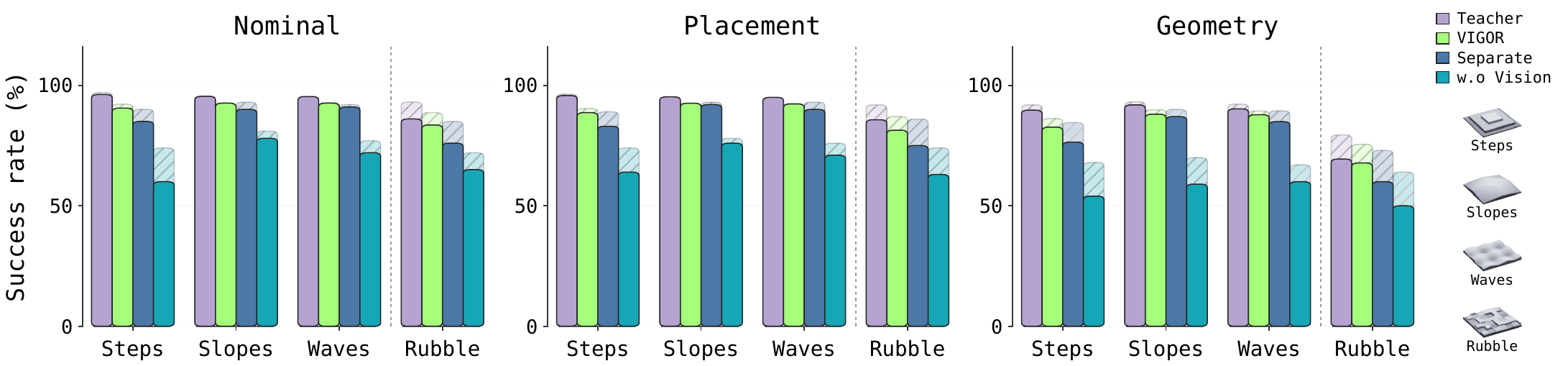}}
\endgroup
\par\vspace{-1.5mm}
\setlength{\abovecaptionskip}{3pt}
\caption{\added{Recovery across terrain conditions. Solid bars show} SSR\added{;} \hatchsym{} \added{bars show SR (\%). Dividers mark unseen rubble, excluded from training.}}
\label{fig:teacher_generalization}
\vspace{-3mm}
\end{figure*}

\textbf{Baselines.}
We compare \algo with pretrained HOST~\cite{HuangT-RSS-25} and FIRM~\cite{xu2025unified} policies, using HOST's official checkpoints and a FIRM checkpoint provided by its authors. HOST learns separate RL stand-up policies for different initial configurations, while FIRM \added{retrieves keyframe goal features to condition a diffusion policy.} As no prior visual baseline exists for humanoid fall recovery, we analyze perception through student ablations.

\begin{table}[t]
\setlength{\tabcolsep}{1.8pt}
\caption{Average simulated recovery performance.}
\label{tab:AveRealPerf}
\vspace{-2mm}
\centering
\begingroup
\setlength{\fboxsep}{1pt}
\setlength{\fboxrule}{0.8pt}
\fcolorbox{yellow}{white}{%
\begin{tabular}{@{}l ccc@{\hspace{6pt}}ccc@{}}
\toprule
\multirow{2}{*}{\textbf{Method}}
& \multicolumn{3}{c}{\textbf{Stand-Up}}
& \multicolumn{3}{c}{\textbf{{Fall-Recovery}}} \\
\cmidrule(lr){2-4}\cmidrule(l){5-7}
& SR↑ & SSR↑ & Time↓ & SR↑ & SSR↑ & Time↓ \\
\midrule
HOST~\cite{HuangT-RSS-25}&
17.0\pmval{3.1} & 16.1\pmval{3.1} & 6.9\pmval{1.6} & -- & -- & -- \\
FIRM~\cite{xu2025unified}&
35.3\pmval{4.6} & 30.4\pmval{4.1} & 6.2\pmval{1.5} &
22.1\pmval{6.5} & 19.3\pmval{3.6} & 6.9\pmval{2.1} \\
\algo&
\best{90.8\pmval{1.2}} & \best{88.7\pmval{1.7}} & \best{5.1\pmval{2.2}} &
\best{91.3\pmval{0.9}} & \best{90.5\pmval{0.7}} & \best{5.5\pmval{1.0}} \\
\bottomrule
\end{tabular}%
}
\endgroup

\vspace{1mm}
\vspace{0pt}
\end{table}

\begin{figure}[t]
\centering
\setlength{\abovecaptionskip}{3pt}
\includegraphics[width=\linewidth]{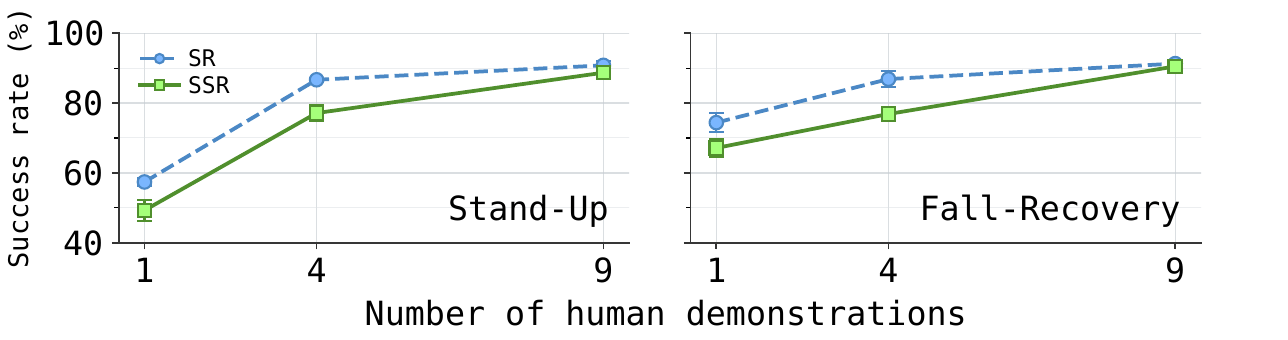}
\vspace{-5mm}
\caption{\added{Demonstration scaling. Recovery success (SR) and stable success (SSR) for visual students supervised by teachers trained with one, four, or nine human demonstrations, under Stand-Up and Fall-Recovery.}}
\vspace{-5mm}
\label{fig:demo_scaling}
\end{figure}

Table~\ref{tab:AveRealPerf} reports average recovery performance across all simulated terrains for both Stand-Up and Fall-Recovery.

\added{VIGOR achieves substantially higher recovery and stable success than the pretrained baselines in the evaluated settings.}

\begin{table}[t]
\setlength{\tabcolsep}{1.8pt}
\caption{Student ablations under Stand-Up and Fall-Recovery.}%
\vspace{-2mm}
\label{tab:StudentAblate}
\centering

\begingroup
\setlength{\fboxsep}{1pt}
\setlength{\fboxrule}{0.8pt}
\fcolorbox{yellow}{white}{%
\begin{tabular}{@{}l ccc@{\hspace{6pt}}ccc@{}}
\toprule
\multirow{2}{*}{\textbf{Method}}
& \multicolumn{3}{c}{\textbf{Stand-Up}}
& \multicolumn{3}{c}{\textbf{Fall-Recovery}} \\
\cmidrule(lr){2-4}\cmidrule(l){5-7}
& SR↑ & SSR↑ & Time↓ & SR↑ & SSR↑ & Time↓ \\
\midrule
w.o Vision&
74.3\pmval{3.2} & 60.9\pmval{1.5} & 5.8\pmval{2.0} &
80.6\pmval{3.0} & 71.2\pmval{1.9} & 5.5\pmval{1.8} \\
w.o Shared&
75.8\pmval{3.7} & 70.9\pmval{2.1} & 6.3\pmval{2.1} &
76.1\pmval{3.2} & 72.5\pmval{2.4} & 6.8\pmval{2.0} \\
w.o History&
86.4\pmval{3.0} & 84.2\pmval{1.8} & 5.1\pmval{1.9} &
89.4\pmval{1.4} & 88.2\pmval{1.5} & \best{5.4\pmval{1.4}} \\
\hl{Separate} & 88.9\pmval{0.8} & 84.3\pmval{1.7} & \best{5.0\pmval{2.1}} & 89.3\pmval{0.7} & 85.3\pmval{0.8} & \best{5.4\pmval{1.1}} \\
\algo&
\best{90.8\pmval{1.2}} & \best{88.7\pmval{1.7}} & 5.1\pmval{2.2} &
\best{91.3\pmval{0.9}} & \best{90.5\pmval{0.7}} & 5.5\pmval{1.0} \\
\bottomrule
\end{tabular}%
}
\endgroup
\vspace{0pt}
\end{table}

\input{paper-template-latex/figures_tex/vision_overlays}

\begin{table}[t]
\setlength{\tabcolsep}{1.8pt}
\caption{\added{Safety and recovery quality.}}
\label{tab:RecoverySafety}
\vspace{-2mm}
\centering
\begingroup
\fontsize{8}{9.5}\selectfont
\setlength{\fboxsep}{1pt}
\setlength{\fboxrule}{0.8pt}
\fcolorbox{yellow}{white}{%
\begin{tabular}{@{}l ccccc@{}}
\toprule
\textbf{Method} & PIF↓ & $F_{\mathrm{c}}$↓ & FR↓ & Drift↓ & MPJPE↓ \\
& & (kN) & (\%) & (m) & (cm) \\
\midrule
w.o Vision & 44.4\pmval{7.7} & 3.3\pmval{1.1} & 22.1\pmval{3.0} & 0.9\pmval{0.6} & 21.5\pmval{6.6} \\
\algo & 40.6\pmval{7.6} & 2.0\pmval{1.0} & 7.3\pmval{1.6} & 0.5\pmval{0.5} & 17.3\pmval{5.9} \\
\midrule
Teacher & 39.3\pmval{6.0} & 1.8\pmval{1.1} & 3.1\pmval{2.5} & 0.5\pmval{0.4} & 11.5\pmval{4.8} \\
\bottomrule
\end{tabular}%
}
\endgroup

\end{table}

\input{paper-template-latex/figures_tex/teaser}

\textbf{Student Ablations.}
We ablate student-side components of \algo\ while keeping training conditions fixed. w.o Shared removes intermediate latent supervision, retaining action imitation. w.o Vision removes egocentric depth input, yielding a proprioception-only student. w.o History removes temporal observation history, restricting the student to single-step observations. \algo\ includes egocentric depth, temporal history, and intermediate latent supervision. \added{Separate retains latent supervision but replaces the shared goal--terrain projection with separate encodings, whose concatenation the student predicts.}

Table~\ref{tab:StudentAblate} \added{shows that vision and intermediate latent supervision both contribute substantially to recovery. Removing vision causes the largest reduction in stable success in both tasks, while removing intermediate latent supervision also substantially degrades recovery, supporting recovery-level guidance beyond action imitation. The blind student inherits strong recovery skills from the teacher and can find support and regain a stable posture even after terrain collisions.} \added{Separate achieves comparable recovery success but lower stable success. An additional evaluation also finds higher peak contact forces for Separate than VIGOR (2.4 versus 2.0\,kN).} \added{Temporal history provides a clearer benefit in Stand-Up, where successive contacts must be coordinated. Its smaller benefit in Fall-Recovery is consistent with greater reliance on immediate feedback during rapid descent.}

\textbf{\added{Recovery Across Terrain Conditions.}}
\label{sec:teacher_terrain_coverage}
\added{Factorization lets flat-ground demonstrations guide recovery across varied terrain contacts. We evaluate the teacher and students on steps, slopes, waves, and unseen rubble (Fig.}~\colorbox{yellow}{\ref{fig:teacher_generalization}}\added{), without new demonstrations or fine-tuning.} \textit{\added{Nominal}} \added{uses random starts on standard terrain;} \textit{\added{Placement}} \added{varies starts and headings around terrain features;} \textit{\added{Geometry}} \added{increases terrain severity by doubling heights and gradients.}

\added{Placement changes have relatively little effect, whereas increased geometry reduces recovery, especially on rubble. The teacher's performance on unseen rubble supports transfer of flat-ground motion priors. On unseen rubble, VIGOR and Separate reach standing at similar rates, but VIGOR more often maintains stable recovery; the blind policy performs worse on both metrics. We hypothesize that avoiding reconstruction of a separate terrain latent helps VIGOR maintain recovery on unfamiliar terrain.}

\textbf{\added{Demonstration Scaling.}}
\label{sec:student_demo_coverage}
\added{We compare visual students supervised by teachers trained with one, four, or the full demonstration set. Evaluation uses identical held-out starts across four fall directions and the tested terrains.}
\added{The gain from one to four demonstrations (Fig.}~\ref{fig:demo_scaling}\added{) suggests that complementary recovery patterns broaden the student's coverage.}

\textbf{\added{Safety and Recovery Quality.}}
\added{We compare the teacher, VIGOR, and the blind student from matched initial states under both tasks.}
\added{Table}~\colorbox{yellow}{\ref{tab:RecoverySafety}} \added{suggests that vision supports more controlled recovery: improved reliability is accompanied by lower contact forces, less drift, and closer motion tracking. The reduction in contact peaks is more pronounced than the change in PIF, highlighting the benefit of vision for controlling robot--terrain loading. The teacher's stronger reliability and tracking indicate further room for improvement in distillation.} \added{Fig.}~\colorbox{yellow}{\ref{fig:vision_overlays}} \added{shows recorded recovery sequences for the visual and blind students.}




\begin{figure}[!t]
\centering

\includegraphics[width=\linewidth,trim=0pt 0pt 0 0,clip]{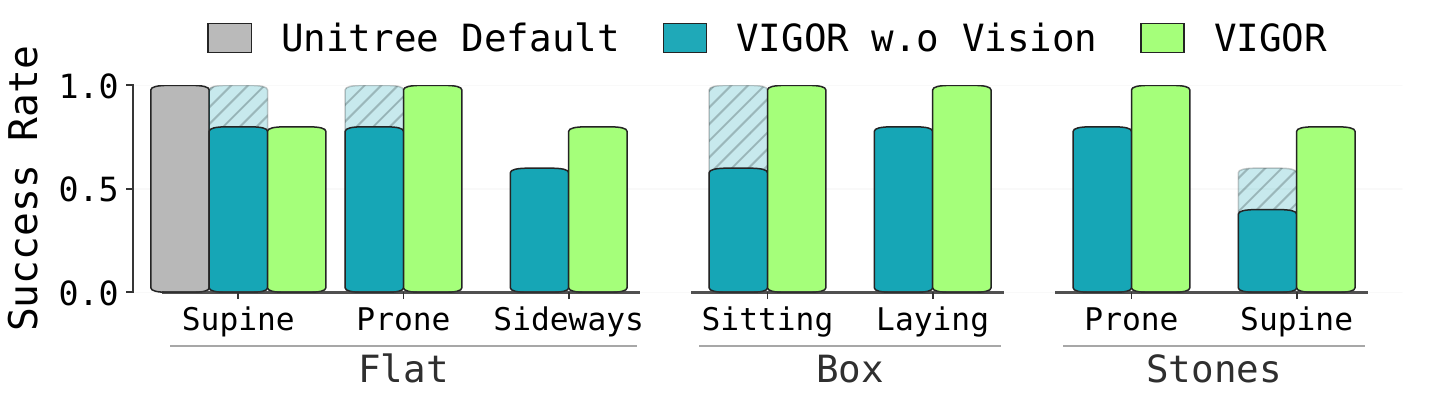}\\
\includegraphics[width=\linewidth,trim=0pt 0pt 0 25,clip]{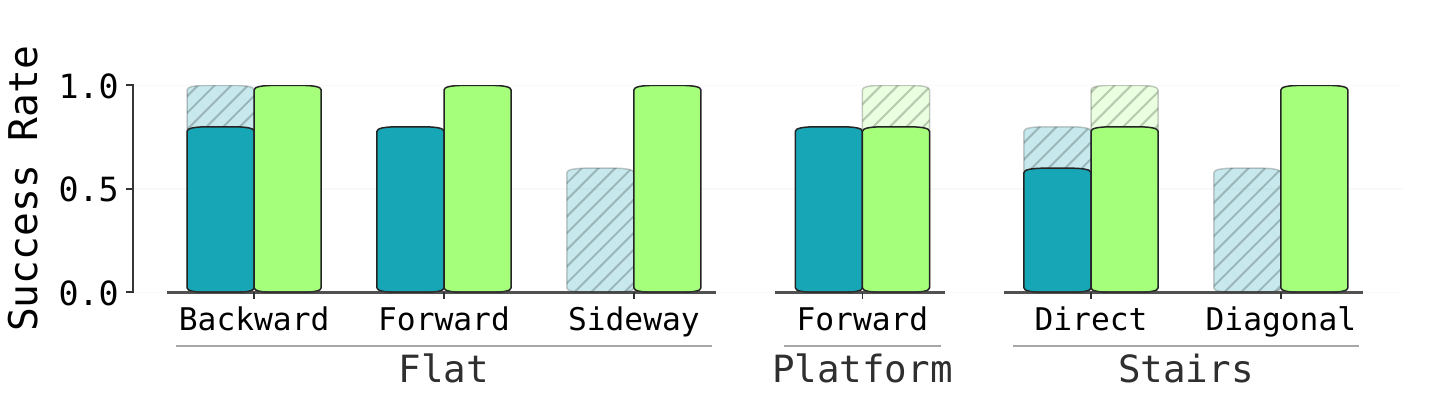}
\vspace{-0.6cm}
\caption{
Real-world recovery performance across surfaces and initial configurations.
Top: stand-up. Bottom: fall recovery. Bars show success over 5 trials; \hatchsym{} denotes successful recoveries with unsafe upper-body contact.
}
\label{fig:RealPerf}
\vspace{-5mm}
\end{figure}

\subsection{Real-World Experiments}

We evaluate \algo on a Unitree~G1 without task-specific tuning or parameter changes. Each setting is tested for 5 trials under the \textit{Stand-Up} and \textit{Fall-Recovery} regimes, \added{with representative sequences shown in Fig.}~\ref{fig:hardware_rollouts}. \added{See the supplementary video for the setups.}

\textbf{Stand-Up.}
We test stand-up on flat ground, stones, and a box obstacle. Flat-ground trials include prone, supine, and sideways initializations; stones include prone and supine initializations; and box trials include seated and torso-on-box configurations. In all cases, joint configurations are \textit{randomized}, and the policy must recover to an upright stance.

\textbf{Fall-Recovery.}
We test fall recovery on flat ground, a raised platform, and stairs. Flat-ground trials use backward, forward, and sideways pushes; platform trials push the robot forward off the edge; and stair trials use direct and diagonal pushes toward the stairs. To ensure a genuine dynamic fall, the policy is initially disabled and activated only after the projected gravity vector indicates that the robot has reached a predefined fall angle ($20^\circ$). \added{This delayed activation evaluates mitigation and recovery after destabilization.}

\textbf{Results.}
Fig.~\ref{fig:RealPerf} reports real-world performance on a Unitree~G1 under \textit{Stand-Up} and \textit{Fall-Recovery}, grouped by terrain and initialization or disturbance type. \added{Fig.}~\ref{fig:hardware_rollouts} \added{shows corresponding recovery behaviors}. \textit{Unitree} controller only succeeds for face-up flat-ground stand-up, so we omit it from further evaluation.
\added{For stand-up, both policies achieve high recovery rates across initial poses on flat terrain.} As terrain complexity increases, especially on stones and box obstacles, the benefit of visual grounding becomes clearer: \algo maintains higher success under uneven contacts, indicating more reliable terrain-aware stabilization.
\added{During fall recovery, both policies recover reliably from forward and backward pushes on flat terrain.} Under more challenging disturbances, including sideways pushes and stair interactions, \algo \added{achieves more recoveries without unsafe upper-body contact}, with the strongest gains in settings that require terrain-aware reactions such as diagonal stair pushes.
\added{These results show why recovery success alone is insufficient to characterize fall safety. In sideways falls and diagonal stair pushes, the blind policy sometimes reaches standing despite unsafe upper-body contacts, whereas VIGOR achieves more recoveries without these contacts. The recovery sequences in Fig.}~\ref{fig:hardware_rollouts} \added{illustrate how depth helps the policy identify nearby support and obstacles before impact and coordinate contacts during recovery.}

%% file: paper-template-latex/figures_tex/vision_overlays.tex
\begin{figure}[t]
\centering
\setlength{\abovecaptionskip}{2pt}
\begingroup
\newcommand{\visionframe}[1]{\filmframe{0.247\linewidth}{#1}}
\newcommand{\annotatedframe}[2]{%
\begin{minipage}[b]{0.247\linewidth}%
\setlength{\parskip}{0pt}\offinterlineskip
\filmrail\par\nointerlineskip
\begin{overpic}[width=\linewidth]{#1}%
\put(50,70){\makebox(0,0){\fontsize{5.2}{6}\selectfont\sffamily #2}}%
\end{overpic}\par\nointerlineskip
\filmrail
\end{minipage}%
}
\newcommand{\policyframe}[2]{%
\begin{minipage}[b]{0.247\linewidth}%
\setlength{\parskip}{0pt}\offinterlineskip
\filmrail\par\nointerlineskip
\begin{overpic}[width=\linewidth]{#1}%
\put(7,39){\makebox(0,0){\rotatebox{90}{\imagelabel{\fontsize{6.8}{7}\selectfont\sffamily\bfseries #2}}}}%
\end{overpic}\par\nointerlineskip
\filmrail
\end{minipage}%
}
\setlength{\fboxsep}{1pt}
\fcolorbox{yellow}{white}{%
\begin{minipage}{\dimexpr\linewidth-2\fboxsep-2\fboxrule\relax}
\offinterlineskip
{\scriptsize Same starting pose\hfill Time $\longrightarrow$}\par\vspace{0.4mm}
\policyframe{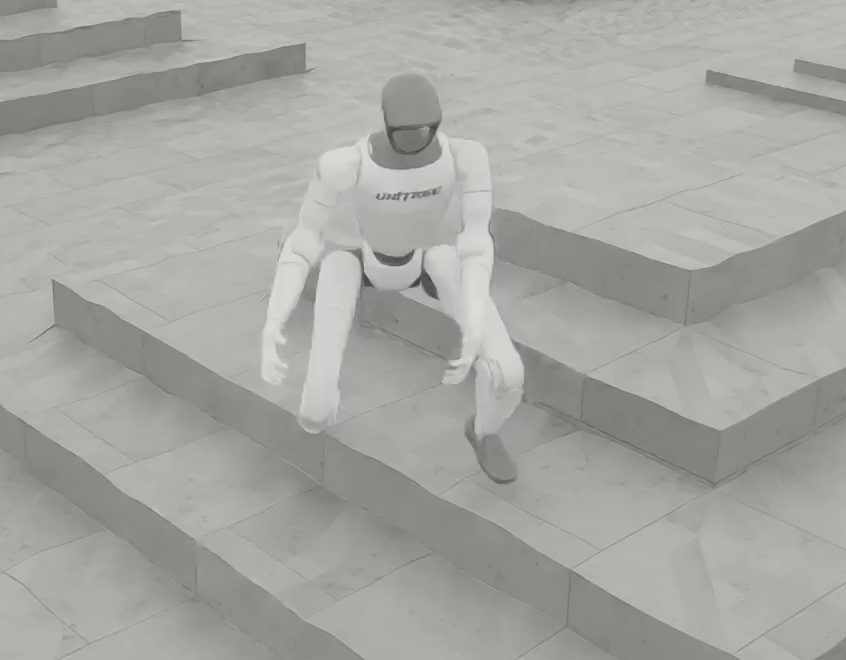}{w.o Vision}\hfill
\annotatedframe{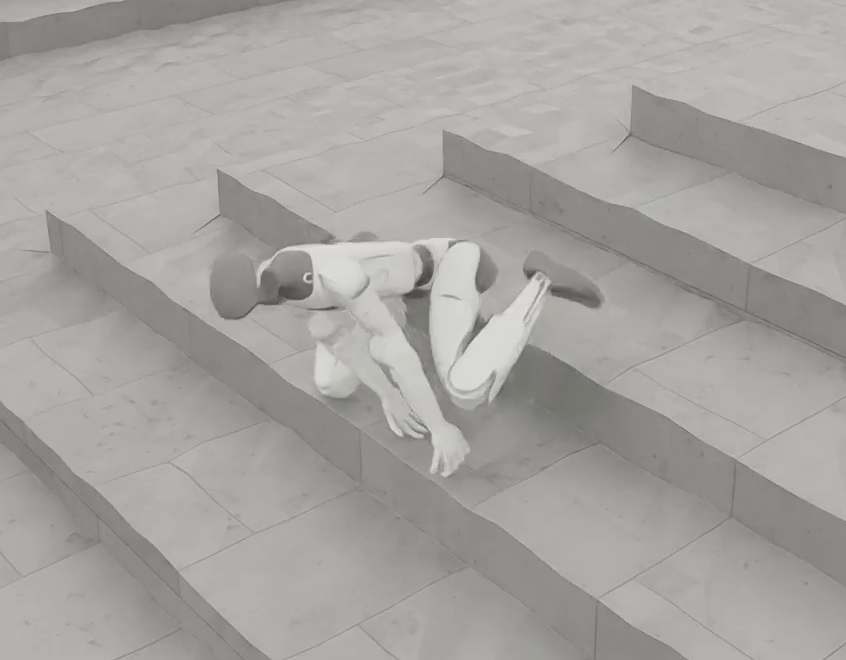}{Leg collision}\hfill
\annotatedframe{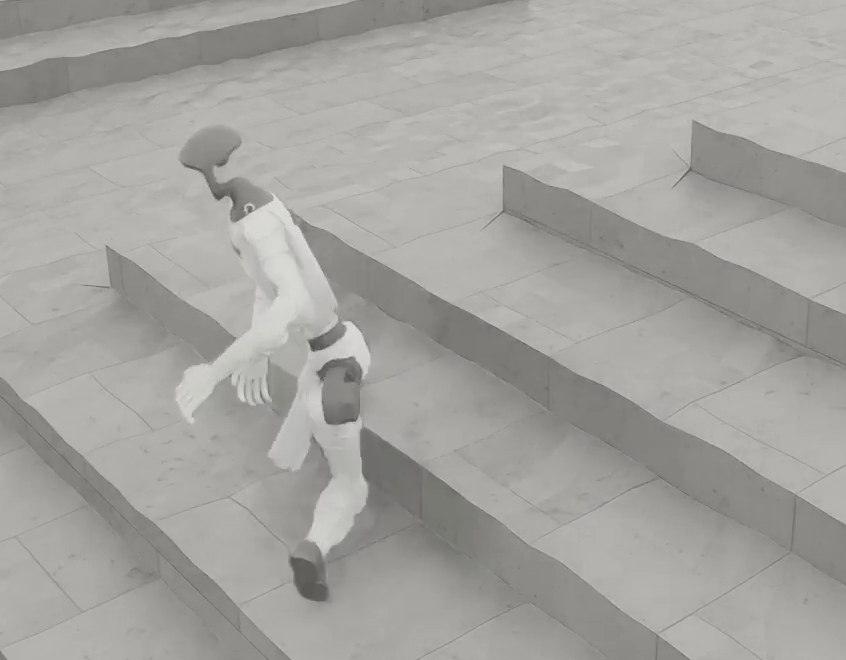}{Awkward posture}\hfill
\visionframe{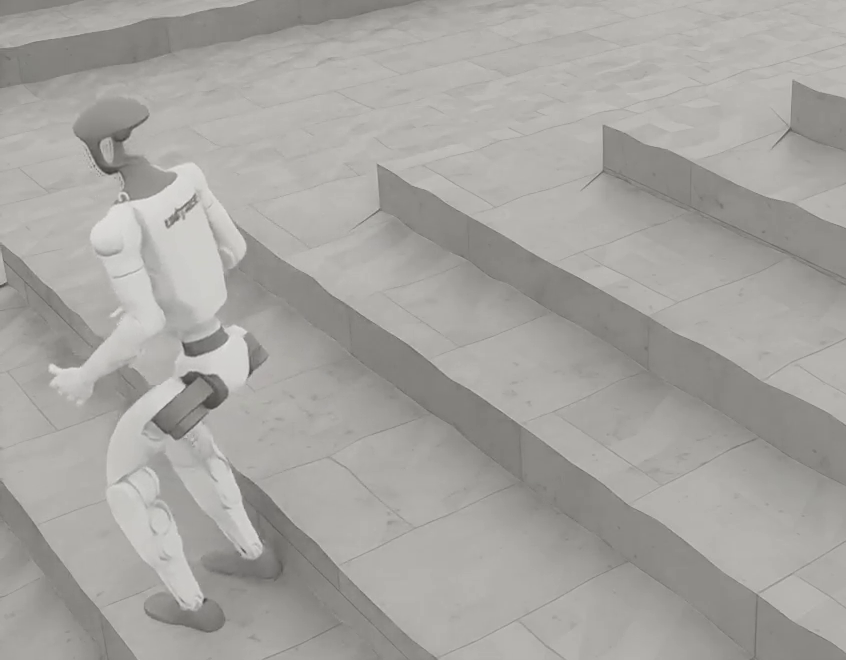}\par\vspace{0.7mm}
\policyframe{figures/vision_filmstrip/blind/001.png}{VIGOR}\hfill
\annotatedframe{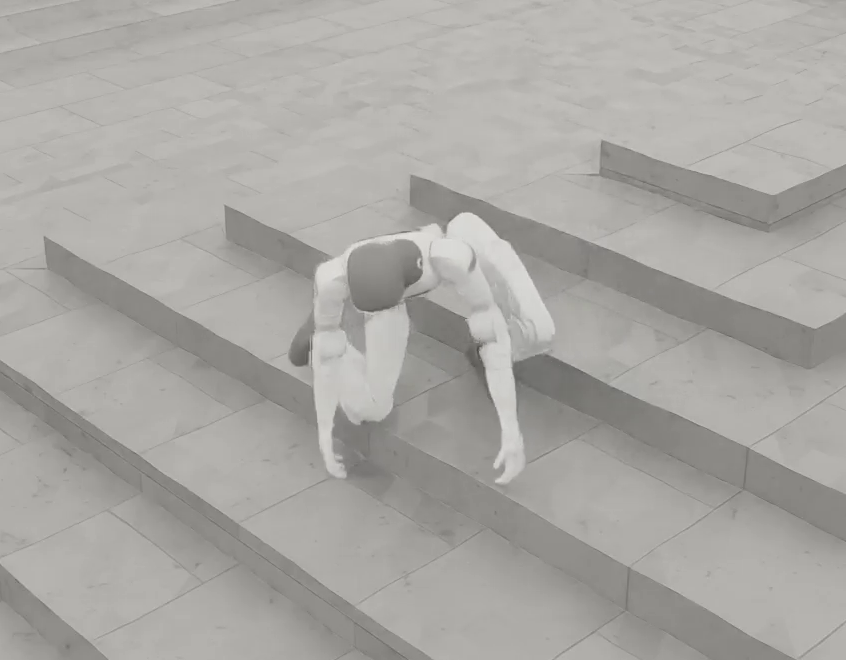}{Protects with hands}\hfill
\visionframe{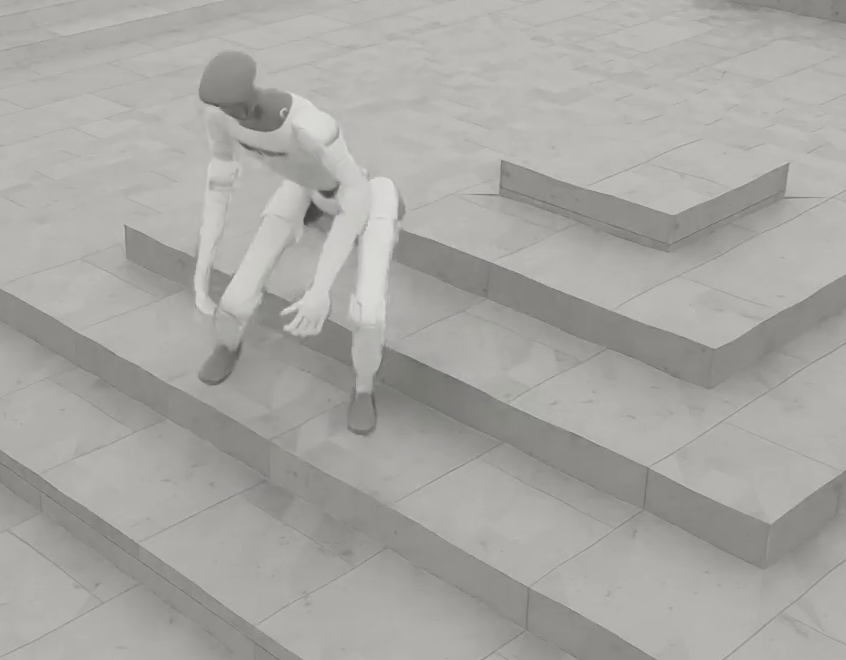}\hfill
\visionframe{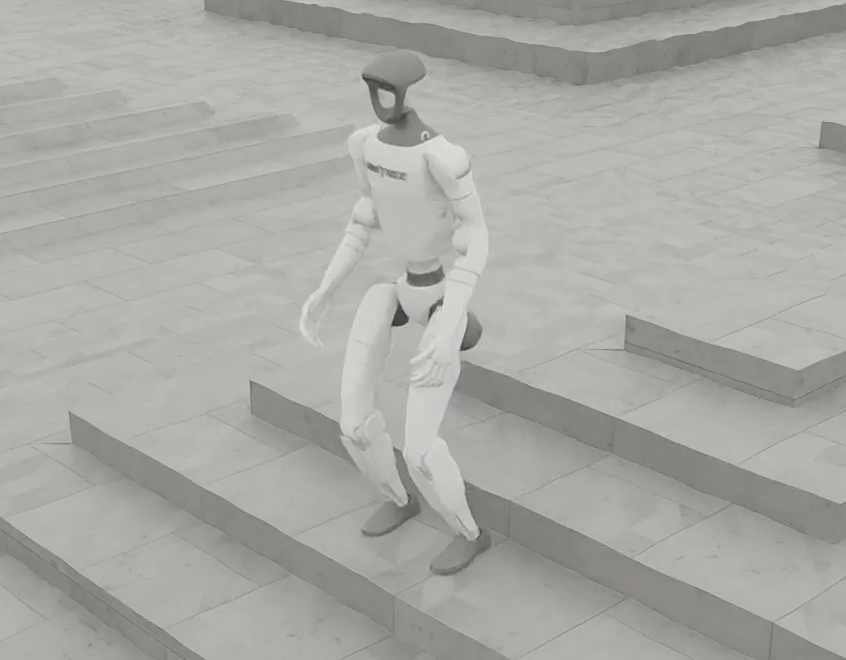}
\end{minipage}%
}
\endgroup
\caption{\added{Stair recovery illustration: w.o Vision (top) and VIGOR (bottom), progressing from left to right.}}
\label{fig:vision_overlays}
\vspace{-4mm}
\end{figure}

%% file: paper-template-latex/figures_tex/teaser.tex
\begin{figure*}[!t]
\centering
\setlength{\abovecaptionskip}{3pt}
\begingroup
\setlength{\fboxsep}{1pt}
\fcolorbox{yellow}{white}{%
\begin{minipage}{\dimexpr\textwidth-2\fboxsep-2\fboxrule\relax}
\begin{minipage}[t]{\linewidth}
\footnotesize\textbf{(A) Diagonal stairs fall}\hfill{\scriptsize Time $\longrightarrow$}\par\vspace{0.4mm}
\begingroup\setlength{\parskip}{0pt}\offinterlineskip
\filmrail\par\nointerlineskip
\begin{overpic}[width=0.247\linewidth,trim=0bp 40bp 0bp 0bp,clip]{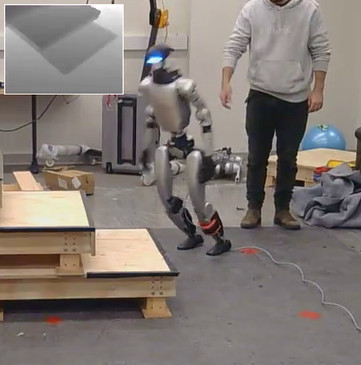}
{\color[HTML]{A5FF7A}\linethickness{1pt}%
\qbezier(74.8,61.7)(67.7,77.9)(57.6,61.7)%
\put(57.6,61.7){\vector(-3,-4){3.03}}%
\put(75.8,76.9){\makebox(0,0){\imagelabel{\scriptsize\sffamily\bfseries Pushed}}}}
\end{overpic}\hfill
\includegraphics[width=0.247\linewidth,trim=0bp 40bp 0bp 0bp,clip]{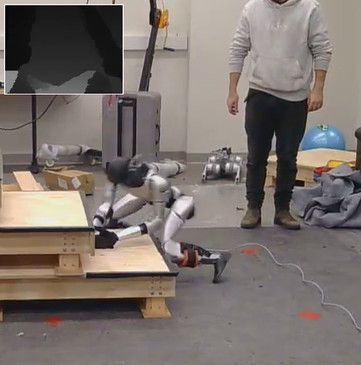}\hfill
\includegraphics[width=0.247\linewidth,trim=0bp 40bp 0bp 0bp,clip]{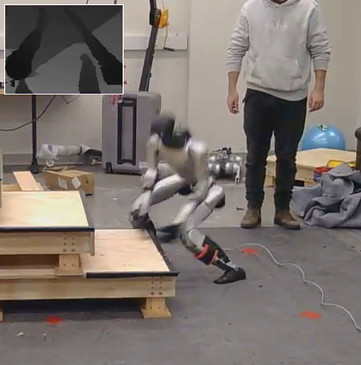}\hfill
\includegraphics[width=0.247\linewidth,trim=0bp 40bp 0bp 0bp,clip]{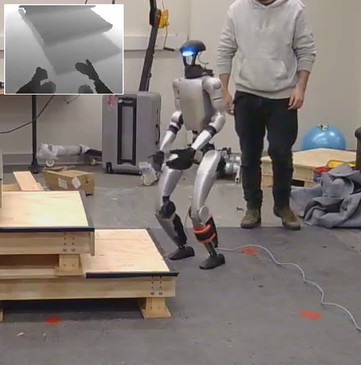}\par\nointerlineskip
\filmrail\par
\endgroup
\end{minipage}\par\vspace{1mm}
\begin{minipage}[t]{0.498\linewidth}
\footnotesize\textbf{(B) Toward stairs}\par\vspace{0.4mm}
\begin{minipage}[b]{0.247\linewidth}
\setlength{\parskip}{0pt}\offinterlineskip
\filmrail\par\nointerlineskip
\begin{overpic}[width=\linewidth]{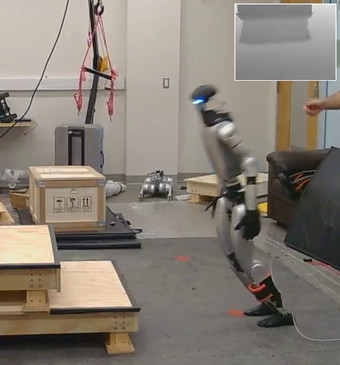}
{\color[HTML]{A5FF7A}\linethickness{0.9pt}%
\qbezier(92,71)(84,87)(78,71)%
\put(78,71){\vector(-1,-2){3}}}
\end{overpic}\par\nointerlineskip
\filmrail
\end{minipage}\hfill
\filmframe{0.247\linewidth}{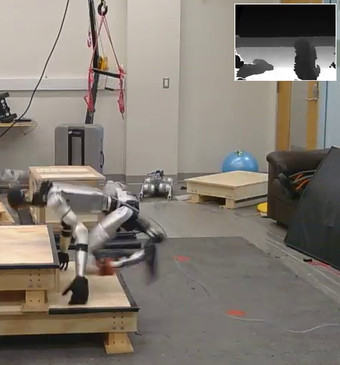}\hfill
\filmframe{0.247\linewidth}{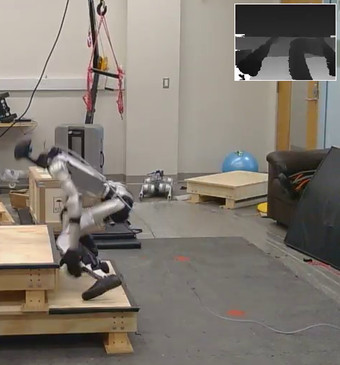}\hfill
\filmframe{0.247\linewidth}{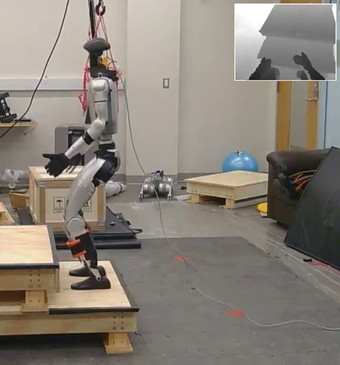}
\end{minipage}\hfill
\begin{minipage}[t]{0.492\linewidth}
\footnotesize\textbf{(C) Legs up (face up)}\par\vspace{0.4mm}
\filmframe{0.247\linewidth}{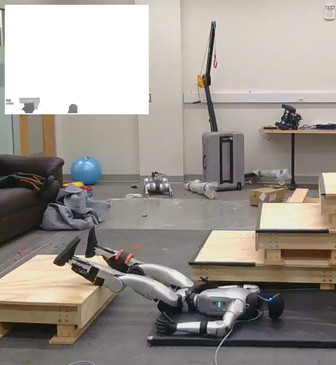}\hfill
\filmframe{0.247\linewidth}{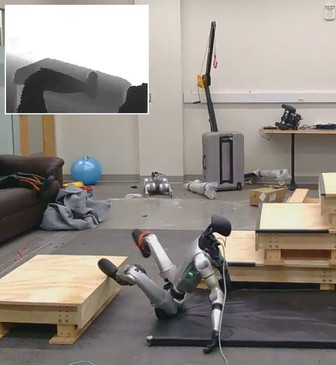}\hfill
\filmframe{0.247\linewidth}{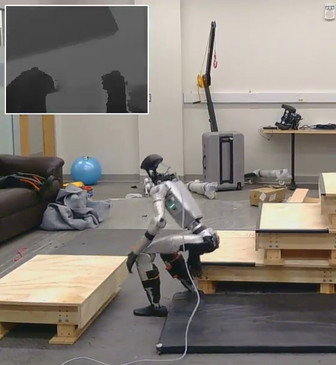}\hfill
\filmframe{0.247\linewidth}{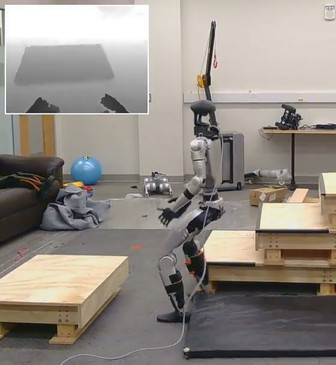}
\end{minipage}\par\vspace{1mm}
\begin{minipage}[t]{0.53\linewidth}
\footnotesize\textbf{(D) Gap (face down)}\par\vspace{0.4mm}
\filmframe{0.247\linewidth}{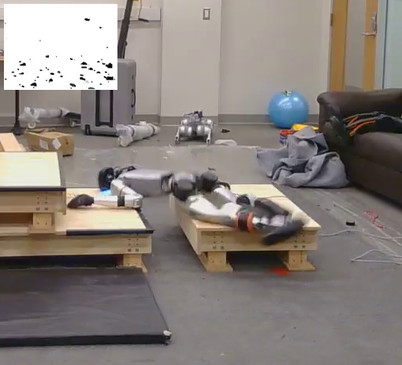}\hfill
\filmframe{0.247\linewidth}{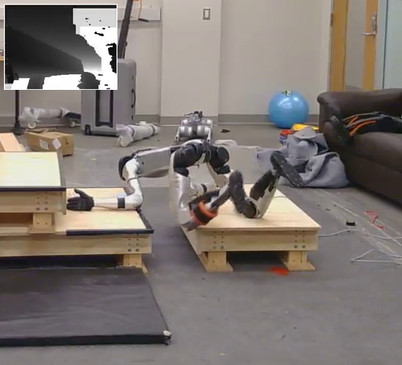}\hfill
\filmframe{0.247\linewidth}{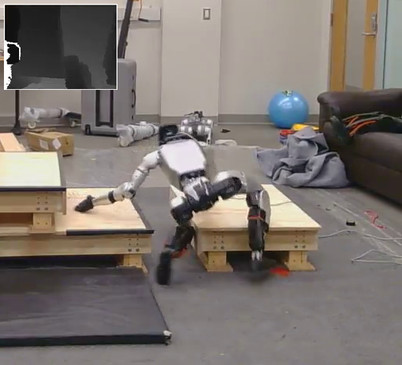}\hfill
\filmframe{0.247\linewidth}{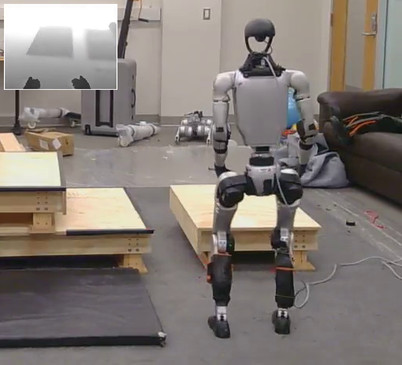}
\end{minipage}\hfill
\begin{minipage}[t]{0.46\linewidth}
\footnotesize\textbf{(E) Gap (face up)}\par\vspace{0.4mm}
\filmframe{0.247\linewidth}{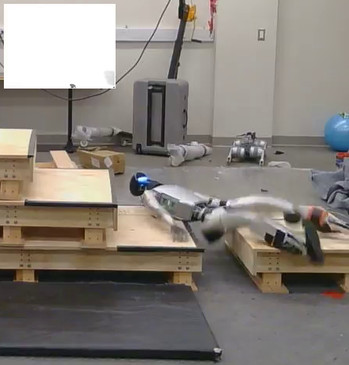}\hfill
\filmframe{0.247\linewidth}{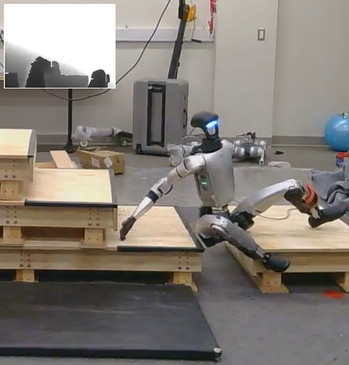}\hfill
\filmframe{0.247\linewidth}{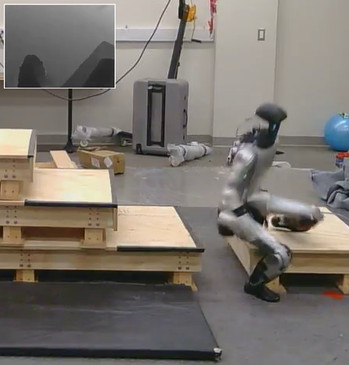}\hfill
\filmframe{0.247\linewidth}{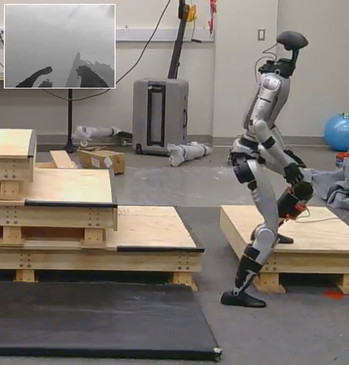}
\end{minipage}\par\vspace{1mm}
\end{minipage}%
}
\endgroup
\caption{\added{Real-world recovery sequences from the project teaser: (A) diagonal stairs fall, (B) fall toward stairs, (C) stand-up with raised legs, and stand-up across a gap from (D) face-down and (E) face-up configurations. Four selected frames per example show the starting pose or fall, contact transitions, and recovery from left to right; insets show synchronized depth.}}
\label{fig:hardware_rollouts}
\vspace{-3mm}
\end{figure*}

%% file: paper-template-latex/5_conclusions.tex
We present \algo, a unified framework for humanoid fall safety that integrates fall mitigation and recovery in a single vision-conditioned policy. By factorizing data into sparse human motion priors and independently varying terrain, and by representing recovery goals through a compact goal-in-context latent, \algo enables RL to resolve contact timing and terrain-dependent execution without dense demonstrations. A privileged teacher learns terrain-aware recovery from sparse keyframes and terrain access, and is distilled into a deployable student operating from egocentric perception. Simulation and real-world results demonstrate robust recovery across diverse terrains and initial conditions, validating structured teacher-student learning as a practical path to zero-shot sim-to-real humanoid fall safety. \added{Future work will explore integration with ongoing tasks, such as locomotion, to support task continuation after recovery.}